\documentclass[10pt, a4paper]{article}
\usepackage{lrec}
\usepackage{graphicx}

\usepackage{balance}
\usepackage{url}
\usepackage[toc,page]{appendix}
\usepackage{diagbox}
\usepackage{amssymb}
\usepackage{multirow}
\usepackage{makecell}
\usepackage[table]{xcolor}

\usepackage{tabularx}
\usepackage{soul}
\usepackage{multirow}

\usepackage{epstopdf}
\usepackage[latin1]{inputenc}

\usepackage{hyperref}
\usepackage{xstring}

\newcommand{\secref}[1]{\StrSubstitute{\getrefnumber{#1}}{.}{ }}

\title{Evaluation of Greek Word Embeddings\\}

\name{Stamatis Outsios$^1$, Christos Karatsalos$^1$, Konstantinos Skianis$^2$, Michalis Vazirgiannis$^{1,2}$}

\address{ $^1$Athens University of Economics and Business, $^2$'Ecole Polytechnique\\
         Greece, France \\
         \{soutsios, ckarats, skianis.konstantinos, michalis.vazirgiannis\}@gmail.com\\}

\abstract{
Since word embeddings have been the most popular input for many NLP tasks, evaluating their quality is critical.
Most research efforts are focusing on English word embeddings. 
This paper addresses the problem of training and evaluating such models for the Greek language. 
We present a new word analogy test set considering the original English Word2vec analogy test set and some specific linguistic aspects of the Greek language as well. 
Moreover, we create a Greek version of WordSim353 test collection for a basic evaluation of word similarities. 
Produced resources are available for download.
We test seven word vector models and our evaluation shows that we are able to create meaningful representations. 
Last, we discover that the morphological complexity of the Greek language and polysemy can influence the quality of the resulting word embeddings. \\ \newline }

\begin{document}

\maketitleabstract

\section{Introduction}

Many unsupervised learning techniques have been proposed to obtain representations of words from text.
Word embeddings, have been found to provide meaningful representations for words in an efficient way, so that they have become very common in many NLP tasks. 


In most cases word embeddings are obtained as a product of training neural network-based models. 
Language modelling is a typical NLP task, where the objective is to predict the probability of the distribution over the ``next" word. 
A word embedding is a vector that has a finite dimension, with the value of each dimension being a feature that weights the relation of the word with a ``latent" aspect of the language. 
These features are jointly learned during unsupervised learning, with plain text data as input, which is not annotated. 
This principle is known as the distributional hypothesis \cite{harris-1954}. The direct implication of this hypothesis is that the word meaning is related to the context where it usually occurs, so it is possible to compare the meanings of two words by applying statistical comparisons of their contexts. 
All these implications were confirmed by empirical tests carried out on human groups in \cite{rubenstein1965contextual,charles2000contextual}.

Word embeddings are produced by performing unsupervised learning, so their evaluation is critical. 
Much has been investigated about word embeddings of English words and phrases, but only little attention has been dedicated to other languages.
The main objective of this work is to explore the behavior of state-of-the-art word embedding methods on Greek, which is a language that is characterized by a very rich morphology. 

We introduce a new test set for the word analogy task that inspects syntactic, morphosyntactic and semantic properties of Greek words and phrases. 
The proposed evaluation scheme is based on word analogies that were presented in \cite{mikolov2013distributed}. 
Our main goal is to compare empirically the performance of different models trained on the largest corpus available so far, collected from about 20M URLs with Greek language content \cite{outsios2018word}. 
Moreover, we introduce a newly created Greek word analogy test set and a basic word similarity resource based on the original dataset WordSim353 \cite{finkelstein2002placing} translated to Greek as well.

These two datasets and also the best word embeddings model according to our evaluation, are available at Greek language resources\footnote{\url{http://archive.aueb.gr:7000/resources/}\label{resource-url}}.

\section{Related Work}

One of the most popular techniques for building distributional models is to train a neural network \cite{mikolov2013distributed} to predict a word given a context (CBOW), or a context given a word (Skip-gram), on the basis of a corpus in which every word occurrence represents one learning example. 
In this approach, word sense is represented as a vector which is actually a row in the neural network's input-to-hidden weight matrix. 
This approach is used for the experiments of this work, to train models having as input Greek text data.

Since word embeddings are produced by performing unsupervised learning, their evaluation is a critical issue.
There is a wealth of research on evaluating word embeddings, which can be broadly divided into intrinsic and extrinsic evaluation methods. 
Intrinsic evaluation mostly relies on analogy questions and measures the similarity of words in a low-dimensional embedding space  \cite{mikolov2013efficient,gao2014wordrep,schnabel2015evaluation}. 
Extrinsic evaluation assesses the quality of the word embeddings as features in models for other tasks, such as part-of-speech tagging \cite{collobert2011natural} and sentiment analysis \cite{schnabel2015evaluation}.  

Most of the proposed evaluation schemes are based on word analogies that were presented in \cite{mikolov2013distributed} for the English language.
For the Arabic language \cite{elrazzaz2017methodical}, a benchmark has been created so that it can be utilized to perform intrinsic evaluation of different word embeddings. 
An evaluation analogy test set has been proposed for Croatian \cite{svoboda2017evaluation}, which consists of semantic analogies and syntactic analogies as in the original one presented in \cite{mikolov2013distributed} for the English language. 
Moreover, research on the evaluation of word embeddings has been published for the Polish language \cite{Mykowiecka2017TestingWE} as well as Czech \cite{svoboda2016new}. 
To the best of our knowledge, no work has been done so far on the evaluation of word embeddings for Greek language.

\section{Greek Web Corpus}
Recently, the largest so far corpus available, crawled from about 20M URLs with Greek language content was presented \cite{outsios2018word}.
After preprocessing, cleaning of the raw crawled text (10TB) and per domain de-duplication (resulted in reducing the size of raw text corpus by 75\%), the final corpus in text form, sized around 50GB. 
Some statistics of the corpus are: 
\begin{itemize}
\setlength\itemsep{0.0em}
\item{\it raw crawled text size}: 10TB
\item{\it final text size}: 50GB 
\item{\it number of tokens}: 3B 
\item{\it number of unique sentences}: 120M
\item{\it number of unigrams}: 7M 
\item{\it number of bigrams}: 90M
\item{\it number of trigrams}: 300M
\end{itemize}
Using this corpus, we produced 5 word embedding models. An interaction with the Greek word vectors is also available at a live web tool\footnote{\url{http://archive.aueb.gr:7000}\label{demo-url}}, offering ``analogy", ``similarity score" and ``most similar words" functions.

\section{Evaluation Framework}
For the evaluation of word embeddings models, mainly two different types of evaluation methods are used, intrinsic and extrinsic.
In this work, we focus on intrinsic evaluation and particularly in word analogy and word similarity tasks.

\subsection{Word Analogy Evaluation}
The first work for capturing relations between words as the offset of their vectors, was presented in \cite{mikolov2013distributed}. 
Solving word analogies is based on the assumption that linear relations between word pairs are indicative of the quality of the embeddings. 

Given a set of three words, $a$, $b$ and $c$, the task is to identify such word $d$ that the relation $c:d$ is the same as the relation $a:b$ \cite{turian2010word,pereira2016comparative}. 

Considering 3CosAdd method \cite{mikolov2013distributed} for solving an analogy $(a-b; c-d)$, we make the computation on the vector representations of words: $(c+b)-a$ and it is expected that the nearest vector to the resulting one, would represent the word $d$. 
For finding the most similar to the resulting one, we use the cosine similarity measure.
An alternative method called 3CosMul was introduced by \cite{levy2014linguistic} for achieving better balance among the different aspects of similarity. 
3CosAdd allows one large similarity term to dominate the expression, while 3CosMul amplifies the differences between small quantities and reduces the differences between larger ones.

We show the full Greek word analogy dataset in Table \ref{tab:greek_analogy}. For a successful prediction of the requested word $d$ in our experiments, we consider the top-1 and also top-5 nearest vectors.

\subsection{Word Similarity}
Word similarity method is based on the idea that the distances between words in an embedding space could be evaluated through human judgments on the actual semantic distances between them.
For example considering the English words, the distance between \textit{cup} and \textit{mug} defined in a continuous interval [0,1] would be 0.8, since these words are near, but not exact synonymous. 
A human assess the degree of similarity of given word pairs. 
 
Word similarity in a specified vector-space can be obtained by computing the cosine similarity between word vectors of a pair of words in order to be able to obtain the lists of pairs of words sorted according to vector-space similarity and human assessment similarity. 
The more similar they are, the better the embeddings are \cite{baroni2014don}.
Computing Spearman's correlation \cite{well2003research} between these ranked lists provides insight into how well the learned word vectors capture intuitive notions of word similarity.

\section{Proposed Resources}
In this section we introduce a newly created Greek word analogy test set and a basic word similarity data-set.
\subsection{Proposed Analogy Questions}
Our Greek analogy test set contains 39,174 questions divided into semantic and syntactic analogy questions. It is larger than the original one for the English language \cite{mikolov2013efficient} and is enhanced with specific linguistic aspects of the Greek language.

The syntactic questions are evaluating the grammatical structure of the word representations, while the semantic are evaluating the capture of the meaning of the vocabulary terms arranged with the grammatical structure. Moreover, it is critical to mention that morphology and syntax are related in a way that each can affect the other. Hence, the syntactic analogies evaluate the word embeddings considering mainly characteristics like
the part of speech which can influence many morphological properties of a word (e.g. suffixes in the conjugation of a verb etc.).

Greek language has a higher morphological complexity than other languages like English. The conjugations of verbs and the declensions of nouns are much different and more complex in Greek language than in English. Particularly, among different conjugations and declensions of a noun or an adjective, a word has different endings. Moreover, the adjectives in Greek language include different forms depending on the gender. So, to capture this morphological complexity we have developed in our Analogy set more syntactic categories than the ones presented in Mikolov's work \cite{mikolov2013efficient}.

The final semantic and syntactic analogy questions for each category include word pairs that are produced by computing all the unique ordered arrangements of two different word pairs. 
All analogy questions that include the target word more than once in the same 4-tuple, have been excluded from the final analogy dataset, just to prevent performance drop for all models. 

\begin{table}[ht]
\centering
\begin{tabular}{|l|r|r|}
\hline
\multicolumn{1}{|c|}{Relation} & \#pairs & \#tuples \\
\hline
\multicolumn{3}{|c|}{Semantic: (13650 tuples)}\\
\hline
common\_capital\_country & 42 & 1722 \\
all\_capital\_country & 78 & 6006 \\
eu\_city\_country & 50 & 2366 \\
city\_in\_region & 40 & 1536 \\
currency\_country & 24 & 552 \\
man\_woman\_family & 18 & 306 \\
profession\_placeof\_work & 16 & 240 \\
performer\_action & 24 & 552 \\
politician\_country & 20 & 370 \\
\hline
\multicolumn{3}{|c|}{Syntactic: (25524 tuples)} \\
\hline
 man\_woman\_job & 26 & 650 \\
 adjective\_adverb & 28 & 756 \\
 opposite & 35 & 1190 \\
 comparative & 36 & 1260 \\
 superlative & 25 & 600 \\
 present\_participle\_active & 48 & 2256 \\
 present\_participle\_passive & 44 & 1892 \\
 nationality\_adjective\_man & 56 & 3080 \\
 nationality\_adjective\_woman & 42 & 1722 \\
 past\_tense  & 34 & 1122 \\
 plural\_nouns & 72 & 5112 \\
 plural\_verbs & 37 & 1332 \\
 adjectives\_antonyms & 50 & 2450 \\
 verbs\_antonyms & 20 & 380 \\
 verbs\_i\_you & 42 & 1722 \\
 \hline
\end{tabular}

\caption{The Greek word analogy test set.}\label{tab:greek_analogy}
\end{table}

Semantic questions are divided into 9 categories and include 13,650 questions in total. Their categories are the following:

\begin{itemize}

\item {\it common\_capital\_country}: 42 most common countries and their corresponding capital cities.
This group of word pairs has been extended having almost two times higher number of word pairs in comparison with the original one for English.

\item{\it all\_capital\_country}: 78 word pairs. 
It includes more pairs than the common ones of the previous category.

\item{\it eu\_city\_country}: 50 word pairs of E.U. cities and their corresponding countries. 
This category is not included in the original Word2Vec analogies, although the interpretation is similar to the U.S. city-state category of the original Word2Vec analogies.

\item{\it city\_in\_region}:  40 word pairs of Greek cities and their corresponding regions.
This category is related to the geographical and administrative structure of the Greek State.

\item{\it currency\_country}: 24 word pairs of type currency-country having the currency name written in full.

\item{\it man\_woman\_family}: 18 word pairs with family relation.

\item{\it profession\_placeof\_work}: 16 word pairs of various professions and their corresponding places of work.

\item{\it performer\_action}: 24 word pairs of type performer-action.

\item{\it politician\_country}: 20 word pairs of politicians and their corresponding countries.
This category as well the two previous ones are not included in the original Word2Vec analogies.
\end{itemize}

To sum up, we have extended the semantic categories considering more aspects of the semantic properties of the language.
The last three categories could also be a part of the analogy test set for other languages apart from the Greek, because they are not dependent on the Greek language particularly. 

Syntactic questions are divided into 15 categories, which are mostly language specific. They include 25,524 questions and are divided in the following categories:

\begin{itemize}
\item{\it man\_woman\_job}: 26 pairs of different professions in masculine-feminine form.

\item{\it adjective\_adverb}: 28 pairs of adjectives and their representatives in adverb form.

\item{\it opposite}: 35 pairs of adjectives and their corresponding opposites. 
The word pairs of this category are not too complex and all of them include the ``privative alpha", which is a Greek prefix that expresses negation or absence of the concept meaning by the term to which it is attached, and it is like the preposition ``un" or ``in" considering the English language.

\item{\it comparative}: 36 word pairs of adjectives (positive form) and their comparative form (i.e. good:better).

\item{\it superlative}: 25 word pairs of adjectives (comparative form) and their corresponding superlative form (i.e. better:best).

\item{\it present\_participle\_active}: 48 pairs of verbs and their corresponding present participles in active voice. 

\item{\it present\_participle\_passive}: 44 pairs of verbs and their corresponding present participles in passive voice. In Greek the present participles in the passive voice have adjective endings.

\item{\it nationality\_adjective\_man}: 56 pairs of countries and their corresponding nationality adjectives in the masculine form. The adjective that represents nationality depends on the gender, in contrast with the relevant characteristic of the English language.

\item{\it nationality\_adjective\_woman}: 42 pairs of countries and their corresponding nationality adjectives in the feminine form.

\item{\it past\_tense}: 34 pairs of verbs and their past tense form.

\item{\it plural\_nouns}: 72 pairs of nouns and their plural form.

\item{\it plural\_verbs}: 37 pairs of verbs in the singular form and their corresponding plural form. 
All words of this category are first person verbs.

\item{\it adjectives\_antonyms}: 50 pairs of antonyms.
The word pairs of this category are more complex than the ones of the category ``opposite" that has been described above.

\item{\it verbs\_antonyms}: 20 pairs of verb antonyms.

\item{\it verbs\_i\_you}: 42 pairs of verbs in the first and second person of singular forms respectively in the singular form.
In Greek, the suffix of a verb in the present tense differs among the different conjugations that it is formed.
\end{itemize}

\subsection{Proposed Word Similarity for Greek}
For performing a basic comparison of the various models of word embeddings considering word similarity, we have also translated in Greek the state-of-the-art English word similarity data-set WordSim353 \cite{finkelstein2002placing}.
This data-set consists of 353 word pairs. Each word pair is manually annotated considering the similarity of the two corresponding words. 
We have excluded five pairs that included acronyms, or it was not feasible to translate an English word into one single Greek word.

\section{Model evaluation}
We explored seven models of Greek word embeddings presented in Table \ref{models}.
In Appendix, Table \ref{models-parameters} we present their train parameters.
Five of these models have been trained on large scale web content \cite{outsios2018word}, mentioned as \textit{gr\_web} in \textit{Corpus} column.

\begin{table}[t]
\centering
\small
\begin{tabular}{c|c|c|c|c}
\hline
	Model & Voc & Tool & Method & Corpus\\ \hline
	gr\_def & 1.3M & fasttext & skip-gram & gr\_web\\ 
	gr\_neg10 & 1.3M & fasttext & skip-gram & gr\_web \\ 
	cc.el.300 & 2M & fasttext & skip-gram & cc+wiki \\ 
	wiki.el & 306K & fasttext & skip-gram & wiki \\ 
	gr\_cbow\_def & 1.3M & fasttext & cbow & gr\_web \\ 
	gr\_d300\_nosub & 1.3M & fasttext & skip-gram & gr\_web \\ 
	gr\_w2v\_sg\_n5 & 1.M & w2v & skip-gram & gr\_web \\ \hline
\end{tabular}
\caption{Evaluated models.\label{models}}
\end{table}

One model has been trained on Wikipedia data using FastText \cite{bojanowski2017enriching}, mentioned as \textit{wiki}.
The last model has been trained on Common Crawl and Wikipedia data using fastText based on CBOW model with position-weights \cite{grave2018learning}, mentioned as \textit{cc+wiki}.

We compared these models in word analogy task. 
Due to space limitations, we show summarized results only for 3CosAdd in Table \ref{top-1} and move the rest in supplementary material.
Considering the two aggregated categories of syntactic and semantic word analogies respectively and both 3CosAdd and 3CosMul metrics, model \textit{cc.el.300} has outperformed all the other models apart from the case of the Syntactic category when we included the \textbf{out-of-vocabulary} (oov) terms, the model \textit{gr\_def} had the best performance.
Model \textit{cc.el.300} was the only one that was trained with CBOW and position-weights.
Model \textit{wiki.el}, trained only on Wikipedia, was the worst significantly almost in every category (and sub-category).
The five models that were trained on the large scale web content \cite{outsios2018word} had lower percentage of oov terms in comparison with the other two.
In some cases where oov terms were considered, they outperformed model \textit{cc.el.300} or had a better ranking considering the accuracy rate in most categories or sub-categories.

In the basic categories, syntactic and semantic, model \textit{gr\_cbow\_def} was the only one that performed much worse in the semantic category than in the syntactic one.
All the other models did not have large differences in performance between semantic and syntactic categories.
In sub-categories, the major factors that had a negative impact on the performance were the high percentage of oov terms and polysemy \cite{gladkova2016intrinsic}. 
We noticed that the sub-category in which most models had the worst performance was \textit{currency\_country} category, again due to polysemy.
Most words that are actual currencies seemed to be repeated for many different countries. 
For example, euro is the currency for Greece, Spain etc and dollar is the currency of America and Canada. 
Sub-categories as \textit{adjectives\_antonyms} and \textit{performer\_action} had the highest percentage of out-of-vocabulary terms, so we observed lower performance in these categories for all models (Table \ref{top-1-sim}).

Moreover, there are findings related to the morphological complexity of the Greek language. We observe that the performance in the category of the \textit{nationality\_adjectives} for masculine is significantly better than the one for the feminine in all models and using all metrics (3CosAdd or 3Cosmul and top-1 or top-5 similar). Also we observe significantly higher performance in the category \textit{opposite} than in the similar category of antonyms in all models and metrics. The explanation for this performance is that the category of opposites includes simpler pair of words based on the primitive alpha at the beginning of an opposite word. On the other hand, the category of antonyms includes pairs of words that their relationship as antonyms includes higher complexity.

For comparison in the word analogy task, we used both 3CosAdd and 3CosMul methods. 
3CosAdd is a linear sum, that allows one large similarity term to dominate the expression. 
It ignores that each term reflects a different aspect of similarity, and that the different aspects have different scales. 
On the other hand, 3CosMul method amplifies the differences between small quantities and reduces the differences between larger ones. 
Results using 3CosMul method instead of 3CosAdd were slightly better in most cases.

\begin{table*}[t]
\small
\centering
\begin{tabular}{  l | r | c | c | c | c | c | c | c }
\hline
\multicolumn{2}{c|}{Category} & gr\_def & gr\_neg10 & cc.el.300 & wiki.el & gr\_cbow\_def & gr\_d300\_nosub & gr\_w2v\_sg\_n5 \\ \hline
Semantic & no oov words & 58.42\% & 59.33\% & \bf 68.80\% & 27.20\% & 31.76\% & 60.79\% & 52.70\% \\
 & with oov words & 52.97\% & 55.33\% & \bf 64.34\% & 25.73\% & 28.80\% & 55.11\% & 47.82\% \\\hline
Syntactic & no oov words & 65.73\% & 61.02\% & \bf 69.35\% & 40.90\% & 64.02\% & 53.69\% & 52.60\% \\
 & with oov words & \bf 53.95\% & 48.69\% & 49.43\% & 28.42\% & 52.54\% & 44.06\% & 43.13\% \\\hline
Overall & no oov words & 63.02\% & 59.96\% & \bf 68.97\% & 36.45\% & 52.04\% & 56.30\% & 52.66\% \\
 & with oov words & 53.60\% & 51.00\% & \bf 54.60\% & 27.50\% & 44.30\% & 47.90\% & 44.80\% \\
 \hline
\end{tabular}
\caption{Summary for 3CosAdd and top-1 nearest vectors.\label{top-1}}
\end{table*}

\begin{table}[ht]
\centering
\resizebox{0.5\textwidth}{!}{
\begin{tabular}{| c | c | c | c |}
\hline
	Model & Pearson & p-value & Pairs (unknown) \\ \hline
	gr\_def & \bf 0.6042 & 3.1E-35 & 2.3\% \\
	gr\_neg10 & 0.5973 & 2.9E-34 & 2.3\% \\
	cc.el.300 & 0.5311 & 1.7E-25 & 4.9\% \\
	wiki.el & 0.5812 & 2.2E-31 & 4.5\% \\
	gr\_cbow\_def & 0.5232 & 2.7E-25 & 2.3\% \\ 
	gr\_d300\_nosub & 0.5889 & 3.8E-33 & 2.3\% \\
	gr\_w2v\_sg\_n5 & 0.5879 & 4.4E-33 & 2.3\% \\ \hline
\end{tabular}
}
\caption{Word similarity.\label{wordsim}}
\end{table}

Considering word similarity, using the Greek version of WordSim353, the percentage for oov terms is low in every case (Table \ref{wordsim}). According to Pearson correlation, \textit{gr\_def} model had the highest correlation with human ratings of similarity.

An extended experimental analysis can be found in the Appendix where we first summary evaluation results for top-1 and top-5 nearest vectors using both 3CosAdd and 3CosMul methods and then provide extended evaluation results per analogy category for all models.

\section{Conclusion}
In this paper we provided an intrinsic evaluation framework for Greek word embeddings, considering the word analogy and similarity tasks.
Moreover, a newly introduced corpus\textsuperscript{\ref{demo-url}} was used for training of the Greek models.

Some of the specific linguistic aspects of the Greek language were added in the word analogy questions\textsuperscript{\ref{resource-url}} test set, which was also introduced as well. WordSim353 data-set was also translated from English to Greek.
We compared seven models in total, five of them trained on large scale web content and showed that our models were able to create meaningful word representations.

As future work we would like to evaluate the Greek word embeddings in other extrinsic tasks like POS tagging, language modelling, text classification.



\section{Bibliographical References}
\bibliography{lrec}
\bibliographystyle{lrec}
\balance

\onecolumn
\appendix
\section{Appendix}
\label{appendix}

\begin{table}[!ht]
\centering
\resizebox{1.0\textwidth}{!}{
\begin{tabular}{| l | l || l | l | l | l | l | l | l |}
\hline

\multicolumn{2}{|c||}{\diagbox{Parameters}{Models}} & gr\_def & gr\_neg10 & cc.el.300 & wiki.el.vec & gr\_cbow\_def & gr\_d300\_nosub & gr\_w2v\_sg\_n5 \\ \hline
	\multicolumn{2}{|c||}{Corpus} & gr\_web & gr\_web & cc+wiki & wiki & gr\_web & gr\_web & gr\_web \\ \hline
	\multicolumn{2}{|c||}{Tool} & fasttext & fasttext & fasttext & fasttext & fasttext & fasttext & word2vec \\ \hline
	\multicolumn{2}{|c||}{Method} & skip-gram & skip-gram & skip-gram & skip-gram & cbow & skip-gram & skip-gram \\ \hline
	-minCount & minimal number of word occurrences & 11 & 11 & 5 & 5 & 11 & 11 & 11 \\ \hline
	-minn & min length of char ngram & 3 & 5 & 3 & 3 & 3 & 0 & 0 \\ \hline
	-maxn & max length of char ngram & 6 & 5 & 6 & 6 & 6 & 0 & 0 \\ \hline
	-dim & size of word vectors & 300 & 300 & 300 & 300 & 300 & 300 & 300 \\ \hline
	-ws & size of the context window & 5 & 5 & 5 & 5 & 5 & 5 & 5 \\ \hline
	-neg & number of negatives sampled & 5 & 10 & 10 & 5 & 5 & 5 & 5 \\ \hline
	-loss & loss function {ns, hs, softmax} & ns & ns & ns & ns & ns & ns & ns \\ \hline

\end{tabular}
}
\caption{Models and train parameters.\label{models-parameters}}
\end{table}

\begin{table}[!ht]
\centering
\resizebox{1.0\textwidth}{!}{
\begin{tabular}{ l | r || c | c | c | c | c | c | c }
\hline
\multicolumn{2}{c||}{Category} & gr\_def & gr\_neg10 & cc.el.300 & wiki.el & gr\_cbow\_def & gr\_d300\_nosub & gr\_w2v\_sg\_n5 \\
\hline
Semantic & no oov words & 82.65\% & 83.15\% & \bf 88.46\% & 62.57\% & 55.41\% & 82.24\% & 78.19\% \\
 & with oov words & 74.94\% & 75.39\% & \bf 81.68\% & 58.34\% & 50.23\% & 74.56\% & 70.89\% \\\hline
Syntactic & no oov words & 83.67\% & 80.06\% & \bf 85.26\% & 70.91\% & 77.63\% & 75.30\% & 73.72\% \\
 & with oov words & \bf 68.67\% & 65.71\% & 61.55\% & 46.71\% & 63.71\% & 61.80\% & 60.50\% \\\hline
Overall & no oov words & 83.30\% & 81.20\% & \bf 86.60\% & 67.30\% & 69.40\% & 77.90\% & 75.40\% \\
 & with oov words & \bf 70.90\% & 69.10\% & 68.60\% & 50.80\% & 59.00\% & 66.20\% & 64.10\% \\
 \hline
\end{tabular}}
\caption{Summary for 3CosAdd and top-5 nearest vectors.\label{top-5}}
\end{table}

\begin{table}[!ht]
\centering
\resizebox{1.0\columnwidth}{!}{
\begin{tabular}{ l | r || c | c | c | c | c | c | c }
\hline
\multicolumn{2}{c||}{Category} & gr\_def & gr\_neg10 & cc.el.300 & wiki.el & gr\_cbow\_def & gr\_d300\_nosub & gr\_w2v\_sg\_n5 \\
\hline
Semantic & no oov words & 83.72\% & 84.38\% & \bf 88.50\% & 65.85\% & 52.05\% & 83.26\% & 80.00\% \\
 & with oov words & 75.90\% & 76.50\% & \bf 81.70\%	& 61.40\%& 47.20\% & 75.50\% & 72.50\% \\\hline
Syntactic & no oov words & 83.86\% &	 80.42\% & \bf 85.07\% & 72.56\% & 76.22\% & 75.97\% & 74.55\% \\
 & with oov words & \bf 68.80\% & 66.00\% & 61.40\% & 47.80\% & 62.60\% & 62.30\% & 61.20\% \\\hline
Overall & no oov words & 83.80\% &	81.90\%	& \bf 86.50\% & 69.70\% & 67.20\%	& 78.70\% & 76.60\% \\
 & with oov words & \bf 71.29\%	& 69.66\% & 68.48\% & 52.53\% & 57.20\% & 66.93\% & 65.14\% \\
 \hline
\end{tabular}}
\caption{Summary for 3CosMul and top-5 nearest vectors.\label{cosmul-top-5}}
\end{table}

\begin{table}[!ht]
\begin{center}
\resizebox{1.0\textwidth}{!}{
\begin{tabular}{ l | r || c | c | c | c | c | c | c }
\hline
\multicolumn{2}{c||}{Category} & gr\_def & gr\_neg10 & cc.el.300 & wiki.el & gr\_cbow\_def & gr\_d300\_nosub & gr\_w2v\_sg\_n5 \\
\hline
Semantic & no oov words & 60.60\% & 62.50\% & \bf 70.90\% & 37.50\% & 29.80\% & 62.50\% & 54.60\% \\
 & with oov words & 54.90\%	& 57.00\% & \bf 65.50\% & 35.00\% & 27.10\% & 56.60\% & 49.50\% \\\hline
Syntactic & no oov words & 67.90\% & 62.90\% & \bf 69.60\% & 50.70\% & 63.80\% & 56.50\% & 55.40\% \\
 & with oov words & \bf 55.70\% & 51.30\% & 50.20\% & 33.40\% & 52.30\% & 46.40\% & 45.50\% \\\hline
Overall & no oov words & 65.19\% & 62.66\% & \bf 70.12\% & 45.01\% & 	51.18\% & 58.73\% & 55.10\% \\
 & with oov words & 55.46\% & 53.30\% & \bf 55.54\% & 33.94\% & 43.53\% & 49.96\% & 46.87\% \\
 \hline

\end{tabular}}
\caption{Summary for 3CosMul and top-1 nearest vectors.\label{cosmul-top-1}}
 \end{center}
\end{table}

\begin{table}[!ht]
\centering
\resizebox{0.9\textwidth}{!}{
\begin{tabular}{| c || c | c | c | c | c | c | c |}
\hline
\multirow{2}{4em}{Category} & \multicolumn{7}{c|}{\underline{Models}}\\
 & gr\_def & gr\_neg10 & cc.el.300 & wiki.el & gr\_cbow\_def & gr\_d300\_nosub & gr\_w2v\_sg\_n5 \\
\hline
all\_capital\_country & 61.78\% & 65.26\% & 75.42\% & 25.03\% & 26.50\% & 64.50\% & 52.40\% \\
city\_in\_region & 53.60\% & 53.00\% & 47.70\% & 10.20\% & 32.40\% & 61.30\% & 62.20\% \\
common\_capital\_country & 69.57\% & 73.48\% & 83.72\% & 43.29\% & 39.80\% & 72.10\% & 61.00\% \\
currency\_country & 17.00\% & 19.00\% & 22.90\% & 3.20\% & 7.60\% & 17.00\% & 17.80\% \\
eu\_city\_country & 60.40\% & 64.00\% & 78.30\% & 48.60\% & 33.10\% & 60.40\% & 52.20\% \\
man\_woman\_family & 75.83\% & 76.25\% & 87.62\% & 12.08\% & 75.00\% & 72.10\% & 70.80\% \\
performer\_action & 19.00\% & 16.49\% & 22.53\% & 1.78\% & 26.10\% & 21.60\% & 18.50\% \\
politician\_country & 66.33\% & 62.30\% & 70.80\% & 21.00\% & 52.40\% & 56.00\% & 58.70\% \\
profession\_place\_of\_work & 39.52\% & 38.57\% & 59.89\% & 14.29\% & 53.30\% & 47.60\% & 50.50\% \\\hline
Semantic & 58.42\% & 59.33\% & \textbf{68.80}\% & 27.20\% & 31.76\% & 60.79\% & 52.70\%  \\\hline
adjective\_adverb & 30.34\% & 23.50\% & 34.90\% & 8.17\% & 38.60\% & 21.40\% & 25.80\% \\
adjective\_antonyms & 25.40\% & 23.40\% & 31.20\% & 13.10\% & 20.80\% & 23.20\% & 22.70\% \\
comparative & 88.18\% & 75.49\% & 73.54\% & 58.50\% & 84.90\% & 61.00\% & 62.40\% \\
verbs\_i\_you & 97.05\% & 94.49\% & 98.11\% & 83.33\% & 95.30\% & 90.30\% & 89.60\% \\
man\_woman\_job & 83.98\% & 83.33\% & 73.10\% & 39.18\% & 94.60\% & 77.30\% & 80.70\% \\
nationality\_adjective\_man & 81.20\% & 76.20\% & 83.67\% & 62.34\% & 73.10\% & 62.10\% & 53.10\% \\
nationality\_adjective\_woman & 61.40\% & 51.08\% & 58.40\% & 38.17\% & 45.80\% & 37.50\% & 35.20\% \\
opposite & 36.59\% & 30.65\% & 48.15\% & 24.87\% & 37.50\% & 27.60\% & 25.70\% \\
past\_tense & 83.71\% & 76.52\% & 80.69\% & 3.16\% & 89.60\% & 75.50\% & 79.00\% \\
plural\_nouns & 56.22\% & 50.02\% & 63.92\% & 37.10\% & 46.50\% & 49.40\% & 47.90\% \\
plural\_verbs & 98.50\% & 97.75\% & 99.52\% & 85.54\% & 98.60\% & 95.90\% & 95.60\% \\
present\_participle\_(active) & 82.88\% & 72.15\% & 91.96\% & 67.90\% & 96.90\% & 59.20\% & 63.40\% \\
present\_participle\_(passive) & 44.90\% & 21.50\% & 33.46\% & 43.64\% & 80.80\% & 14.60\% & 16.20\% \\
superlative & 61.11\% & 18.06\% & 58.33\% & 50.00\% & 80.60\% & 9.70\% & 5.60\% \\
verbs\_antonyms & 20.90\% & 13.20\% & 19.90\% & 4.40\% & 11.50\% & 14.30\% & 13.20\% \\\hline
Syntactic & 65.73\% & 61.02\% & \textbf{69.35}\% & 40.90\% & 64.02\% & 53.69\% & 52.60\% \\\hline\hline
Overall & 63.02\% & 59.96\% & \textbf{68.97}\% & 36.45\% & 52.04\% & 56.30\% & 52.66\% \\\hline
\rowcolor{lightgray}Questions with oov words & 14.90\% & 14.90\% & 20.80\% & 24.60\% & 14.90\% & 14.90\% & 14.90\% \\\hline
\end{tabular}
}
\caption{Top-1 sim.}\label{top-1-sim}
\end{table}

\begin{table}[!ht]
\centering
\resizebox{0.9\textwidth}{!}{
\begin{tabular}{| c || c | c | c | c | c | c | c |}
\hline
\multirow{2}{4em}{Category} & \multicolumn{7}{c|}{\underline{Models}}\\
 & gr\_def & gr\_neg10 & cc.el.300 & wiki.el & gr\_cbow\_def & gr\_d300\_nosub & gr\_w2v\_sg\_n5 \\
\hline
all\_capital\_country & 54.10\% & 57.10\% & 69.70\% & 23.10\% & 23.20\% & 56.50\% & 45.80\% \\
city\_in\_region & 53.60\% & 53.00\% & 47.70\% & 10.20\% & 32.40\% & 61.30\% & 62.20\% \\
common\_capital\_country & 66.30\% & 70.00\% & 79.70\% & 41.20\% & 37.90\% & 68.70\% & 58.10\% \\
currency\_country & 10.50\% & 11.80\% & 15.80\% & 2.90\% & 4.70\% & 10.50\% & 11.10\% \\
eu\_city\_country & 57.90\% & 61.70\% & 75.10\% & 46.60\% & 31.70\% & 57.90\% & 50.10\% \\
man\_woman\_family & 59.50\% & 59.80\% & 60.10\% & 9.50\% & 58.80\% & 56.50\% & 55.60\% \\
performer\_action & 19.00\% & 19.20\% & 20.10\% & 3.40\% & 26.10\% & 21.60\% & 18.50\% \\
politician\_country & 59.50\% & 55.90\% & 63.50\% & 17.00\% & 47.00\% & 50.30\% & 52.70\% \\
profession\_place\_of\_work & 34.60\% & 33.80\% & 45.40\% & 10.80\% & 46.70\% & 41.70\% & 44.20\% \\\hline
Semantic & 52.97\% & 55.33\% & \textbf{64.34}\% & 25.73\% & 28.80\% & 55.11\% & 47.82\% \\\hline
adjective\_adverb & 28.20\% & 21.80\% & 32.40\% & 6.50\% & 35.80\% & 19.80\% & 23.90\%  \\
adjective\_antonyms & 23.30\% & 21.50\% & 25.20\% & 11.10\% & 19.20\% & 21.30\% & 20.90\% \\
comparative & 56.80\% & 48.70\% & 37.90\% & 37.70\% & 54.70\% & 39.30\% & 40.20\% \\
verbs\_i\_you & 87.90\% & 85.60\% & 84.40\% & 29.00\% & 86.40\% & 81.80\% & 81.20\% \\
man\_woman\_job & 59.70\% & 59.20\% & 38.50\% & 20.60\% & 67.20\% & 54.90\% & 57.40\% \\
nationality\_adjective\_man & 81.20\% & 76.20\% & 80.70\% & 62.30\% & 73.10\% & 62.10\% & 53.10\% \\
nationality\_adjective\_woman & 33.20\% & 27.60\% & 23.80\% & 27.90\% & 24.70\% & 20.30\% & 19.00\% \\
opposite & 30.50\% & 25.50\% & 26.30\% & 15.80\% & 31.30\% & 23.00\% & 21.40\% \\
past\_tense & 78.80\% & 72.00\% & 62.60\% & 1.10\% & 84.30\% & 71.00\% & 74.30\% \\
plural\_nouns & 54.70\% & 48.60\% & 62.10\% & 35.10\% & 45.20\% & 48.00\% & 46.60\% \\
plural\_verbs & 98.50\% & 97.70\% & 94.10\% & 41.70\% & 98.60\% & 95.90\% & 95.60\% \\
present\_participle\_(active) & 69.50\% & 60.50\% & 57.30\% & 35.80\% & 81.20\% & 49.60\% & 53.20\% \\
present\_participle\_(passive) & 18.90\% & 7.40\% & 4.80\% & 2.30\% & 27.70\% & 5.00\% & 5.50\% \\
superlative & 7.30\% & 2.20\% & 1.20\% & 3.50\% & 9.70\% & 1.20\% & 1.00\% \\
verbs\_antonyms & 10.00\% & 6.30\% & 8.20\% & 1.20\% & 5.50\% & 6.80\% & 6.30\% \\\hline
Syntactic & \textbf{53.95}\% & 48.69\% & 49.43\% & 28.42\% & 52.54\% & 44.06\% & 43.13\% \\\hline\hline
Overall & 53.60\% & 51.00\% & \textbf{54.60}\% & 27.50\% & 44.30\% & 47.90\% & 44.80\% \\ \hline
\end{tabular}
}
\caption{Top-1 sim (with out-of-vocabulary words).}
\end{table}

\begin{table}[!ht]
\centering
\resizebox{0.9\textwidth}{!}{
\begin{tabular}{| c || c | c | c | c | c | c | c |}
\hline
\multirow{2}{4em}{Category} & \multicolumn{7}{c|}{\underline{Models}}\\
 & gr\_def & gr\_neg10 & cc.el.300 & wiki.el & gr\_cbow\_def & gr\_d300\_nosub & gr\_w2v\_sg\_n5 \\
\hline
all\_capital\_country & 86.10\% & 87.10\% & 90.90\% & 65.50\% & 48.60\% & 86.50\% & 79.50\% \\
city\_in\_region & 86.70\% & 88.40\% & 88.90\% & 38.90\% & 61.70\% & 86.50\% & 86.20\% \\
common\_capital\_country & 91.20\% & 90.00\% & 94.40\% & 84.90\% & 64.90\% & 90.90\% & 85.70\% \\
currency\_country & 34.50\% & 38.90\% & 46.80\% & 10.10\% & 22.80\% & 32.50\% & 33.60\% \\
eu\_city\_country & 80.80\% & 82.10\% & 90.70\% & 81.80\% & 59.10\% & 80.00\% & 78.40\% \\
man\_woman\_family & 90.80\% & 91.70\% & 92.90\% & 55.80\% & 87.10\% & 88.30\% & 86.20\% \\
performer\_action & 46.20\% & 43.50\% & 58.30\% & 18.40\% & 45.70\% & 44.20\% & 44.70\% \\
politician\_country & 88.00\% & 81.30\% & 91.30\% & 42.00\% & 70.20\% & 83.10\% & 79.20\% \\
profession\_place\_of\_work & 75.20\% & 72.40\% & 89.60\% & 44.00\% & 84.30\% & 74.30\% & 75.70\% \\\hline
Semantic & 82.65\% & 83.15\% & \textbf{88.46}\% & 62.57\% & 55.41\% & 82.24\% & 78.19\% \\ \hline

adjective\_adverb & 53.30\% & 45.20\% & 51.90\% & 37.80\% & 61.10\% & 44.70\% & 47.00\% \\
adjective\_antonyms & 44.50\% & 42.30\% & 54.20\% & 26.70\% & 43.30\% & 41.80\% & 40.30\% \\
comparative & 98.30\% & 94.70\% & 96.90\% & 80.00\% & 96.70\% & 83.70\% & 83.40\% \\
verbs\_i\_you & 97.80\% & 97.40\% & 100.00\% & 94.20\% & 99.10\% & 97.50\% & 97.50\% \\
man\_woman\_job & 98.70\% & 100.00\% & 95.90\% & 72.20\% & 100.00\% & 98.50\% & 98.70\% \\
nationality\_adjective\_man & 97.20\% & 93.80\% & 98.10\% & 91.10\% & 93.90\% & 81.80\% & 76.50\% \\
nationality\_adjective\_woman & 84.90\% & 77.60\% & 83.50\% & 74.30\% & 67.60\% & 65.70\% & 58.90\% \\
opposite & 57.50\% & 52.60\% & 66.80\% & 42.20\% & 64.20\% & 46.70\% & 44.90\% \\
past\_tense & 99.50\% & 98.40\% & 97.60\% & 27.10\% & 100.00\% & 97.80\% & 99.90\% \\
plural\_nouns & 85.50\% & 82.80\% & 83.70\% & 77.50\% & 57.90\% & 81.50\% & 78.40\% \\
plural\_verbs & 100.00\% & 100.00\% & 100.00\% & 95.80\% & 100.00\% & 100.00\% & 100.00\% \\
present\_participle\_(active) & 95.20\% & 91.10\% & 97.80\% & 88.30\% & 99.80\% & 81.90\% & 84.60\% \\
present\_participle\_(passive) & 64.30\% & 39.70\% & 55.90\% & 52.70\% & 90.80\% & 26.30\% & 28.60\% \\
superlative & 75.00\% & 62.50\% & 91.70\% & 66.70\% & 90.30\% & 34.70\% & 30.60\% \\
verbs\_antonyms & 61.00\% & 59.30\% & 59.00\% & 16.70\% & 52.70\% & 56.60\% & 58.20\% \\\hline
Syntactic & 83.67\% & 80.06\% & \textbf{85.26}\% & 70.91\% & 77.63\% & 75.30\% & 73.72\% \\\hline \hline
Overall & 83.30\% &	81.20\% & \textbf{86.60}\% &	67.30\% & 69.40\% & 77.90\% & 75.40\% \\
\hline
\rowcolor{lightgray}Questions with oov words & 14.90\% & 14.90\% & 20.80\% & 24.60\% & 14.90\% & 14.90\% & 14.90\% \\\hline
\end{tabular}
}
\caption{Top-5 sim.}
\end{table}


\begin{table}[!ht]
\centering
\resizebox{0.9\textwidth}{!}{
\begin{tabular}{| c || c | c | c | c | c | c | c |}
\hline

\multirow{2}{4em}{Category} & \multicolumn{7}{c|}{\underline{Models}}\\
 & gr\_def & gr\_neg10 & cc.el.300 & wiki.el & gr\_cbow\_def & gr\_d300\_nosub & gr\_w2v\_sg\_n5 \\
\hline
all\_capital\_country & 75.40\% & 76.20\% & 84.00\% & 60.50\% & 42.60\% & 75.70\% & 69.60\% \\
city\_in\_region & 86.70\% & 88.40\% & 88.90\% & 38.90\% & 61.70\% & 86.50\% & 86.20\% \\
common\_capital\_country & 86.90\% & 85.70\% & 89.90\% & 80.90\% & 61.80\% & 86.50\% & 81.60\% \\
currency\_country & 21.40\% & 24.10\% & 32.20\% & 9.20\% & 14.10\% & 20.10\% & 20.80\% \\
eu\_city\_country & 77.50\% & 78.70\% & 86.90\% & 78.40\% & 56.60\% & 76.70\% & 75.20\% \\
man\_woman\_family & 71.20\% & 71.90\% & 63.70\% & 43.80\% & 68.30\% & 69.30\% & 67.60\% \\
performer\_action & 46.20\% & 43.50\% & 53.40\% & 16.80\% & 45.70\% & 44.20\% & 44.70\% \\
politician\_country & 78.90\% & 73.00\% & 81.90\% & 34.10\% & 63.00\% & 74.60\% & 71.10\% \\
profession\_place\_of\_work & 65.80\% & 63.30\% & 67.90\% & 33.30\% & 73.80\% & 65.00\% & 66.20\% \\ \hline
Semantic & 74.94\% & 75.39\% & \textbf{81.68} \% & 58.34\% & 50.23\% & 74.56\% & 70.89\% \\ \hline
adjective\_adverb & 49.50\% & 41.90\% & 48.10\% & 30.00\% & 56.70\% & 41.50\% & 43.70\% \\
adjective\_antonyms & 40.90\% & 38.90\% & 43.80\% & 22.60\% & 39.90\% & 38.50\% & 37.10\% \\
comparative & 63.30\% & 61.00\% & 50.00\% & 51.60\% & 62.30\% & 54.00\% & 53.70\% \\
verbs\_i\_you & 88.60\% & 88.30\% & 86.10\% & 32.00\% & 89.80\% & 88.30\% & 88.30\% \\
man\_woman\_job & 70.20\% & 71.10\% & 50.50\% & 38.00\% & 71.10\% & 70.00\% & 70.20\% \\
nationality\_adjective\_man & 97.20\% & 93.80\% & 94.60\% & 91.10\% & 93.90\% & 81.80\% & 76.50\% \\
nationality\_adjective\_woman & 45.90\% & 41.90\% & 34.00\% & 54.40\% & 36.50\% & 35.50\% & 31.80\% \\
opposite & 47.90\% & 43.90\% & 36.50\% & 26.80\% & 53.50\% & 38.90\% & 37.40\% \\
past\_tense & 93.70\% & 92.60\% & 75.70\% & 9.20\% & 94.10\% & 92.10\% & 94.00\% \\
plural\_nouns & 83.10\% & 80.50\% & 81.30\% & 73.20\% & 56.30\% & 79.30\% & 76.30\% \\
plural\_verbs & 100.00\% & 100.00\% & 94.60\% & 46.80\% & 100.00\% & 100.00\% & 100.00\% \\
present\_participle\_(active) & 79.90\% & 76.40\% & 60.90\% & 46.60\% & 83.70\% & 68.70\% & 70.90\% \\
present\_participle\_(passive) & 22.10\% & 13.60\% & 8.00\% & 3.20\% & 31.20\% & 9.00\% & 9.80\% \\
superlative & 9.00\% & 7.50\% & 1.80\% & 4.70\% & 10.80\% & 4.20\% & 3.70\% \\
verbs\_antonyms & 29.20\% & 28.40\% & 24.20\% & 3.90\% & 25.30\% & 27.10\% & 27.90\% \\ \hline
Syntactic & \textbf{68.67}\% & 65.71\% & 61.55\% & 46.71\% & 63.71\% & 61.80\% & 60.50\% \\ \hline \hline
Overall & \textbf{70.90}\% &69.10\%&	68.60\%&	50.80\%&	59.00\%&	66.20\%&	64.10\% \\
\hline
\end{tabular}
}
\caption{Top-5 sim (with out-of-vocabulary words).}
\end{table}

\end{document}